\documentclass{article}

\usepackage{PRIMEarxiv}

\usepackage[noend]{algorithmic}
\usepackage{algorithm,caption}
\algsetup{indent=2em}

\usepackage{xcolor}
\definecolor{darkpastelgreen}{rgb}{0.01, 0.75, 0.24}

\usepackage[utf8]{inputenc} 
\usepackage[T1]{fontenc}	
\usepackage{hyperref}   	
\usepackage{url}        	
\usepackage{booktabs}   	
\usepackage{amsfonts}   	
\usepackage{nicefrac}   	
\usepackage{microtype}  	
\usepackage{lipsum}
\usepackage{fancyhdr}   	
\usepackage{graphicx}   	
\graphicspath{{media/}} 	

\usepackage{amsmath} 
\usepackage{svg}
\usepackage[title]{appendix}
\usepackage{adjustbox}

\title{Detach-ROCKET: Sequential feature selection for time series classification with random convolutional kernels
}

\author{%
\normalfont \begin{tabular}{cc}
\begin{tabular}[t]{c}
\textbf{Gonzalo Uribarri}\\ EECS, Science for Life Laboratory \\ and Digital Futures\\ KTH Royal Institute of Technology\\ Stockholm
\end{tabular} &
\begin{tabular}[t]{c}
\textbf{Federico Barone}\\ Department of Mathematics and Geosciences\\ University of Trieste and AREA Science Park\\ Trieste
\end{tabular} \\ \addlinespace[4ex] 
\begin{tabular}[t]{c}
\textbf{Alessio Ansuini}\\ Research and Technology Institute\\ AREA Science Park\\ Trieste
\end{tabular} &
\begin{tabular}[t]{c}
\textbf{Erik Fransén}\\ EECS, Science for Life Laboratory\\  and Digital Futures\\ KTH Royal Institute of Technology\\ Stockholm
\end{tabular}
\end{tabular}
}

\begin{document}
\maketitle

\begin{abstract}

Time Series Classification (TSC) is essential in fields like medicine, environmental science, and finance, enabling tasks such as disease diagnosis, anomaly detection, and stock price analysis. While machine learning models like Recurrent Neural Networks and InceptionTime are successful in numerous applications, they can face scalability issues due to computational requirements. Recently, ROCKET has emerged as an efficient alternative, achieving state-of-the-art performance and simplifying training by utilizing a large number of randomly generated features from the time series data. However, many of these features are redundant or non-informative, increasing computational load and compromising generalization. Here we introduce Sequential Feature Detachment (SFD) to identify and prune non-essential features in ROCKET-based models, such as ROCKET, MiniRocket, and MultiRocket. SFD estimates feature importance using model coefficients and can handle large feature sets without complex hyperparameter tuning. Testing on the UCR archive shows that SFD can produce models with better test accuracy using only 10\% of the original features. We named these pruned models Detach-ROCKET. We also present an end-to-end procedure for determining an optimal balance between the number of features and model accuracy. On the largest binary UCR dataset, Detach-ROCKET improves test accuracy by 0.6\% while reducing features by 98.9\%. By enabling a significant reduction in model size without sacrificing accuracy, our methodology improves computational efficiency and contributes to model interpretability. We believe that Detach-ROCKET will be a valuable tool for researchers and practitioners working with time series data, who can find a user-friendly implementation of the model at \url{https://github.com/gon-uri/detach_rocket}.

\end{abstract}


\section{Introduction}

Time series classification (TSC) refers to the task of assigning a specific label or category to a given sequence of data points that are ordered in time. Its significance extends across various domains, including both foundational research and practical real-world applications, such as heart disease diagnosis \cite{ebrahimi2020review}, speech recognition \cite{hinton2012deep}, environmental monitoring \cite{sarp2017water, davranche2010wetland} and industrial quality control \cite{mehdiyev2017time,chadha2020bidirectional}. With the ever-increasing availability of sensor data and computing devices, the demand for accurate and efficient time series classification algorithms has grown exponentially. This need becomes particularly critical in the domains of healthcare and wearable devices, where the ability to robustly classify time series data in real time carries immense practical implications.

Machine learning (ML) models have proven to be a highly effective solution to TSC problems in a wide range of applications. Models such as Recurrent Neural Networks (RNNs), HIVE-COTE, and InceptionTime have demonstrated remarkable performance in this domain \cite{yu2019review, ismail2020inceptiontime, vaswani2017attention}. However, the large number of learnable parameters and the complexity of the training procedure in such models can affect their scalability and efficiency, especially in the case of large numbers of instances or longer time series. For example, HIVE-COTE has a high training time complexity that scales as $O\left( n^2 \cdot l^4\right)$, where $n$ is the number of training samples and $l$ is the length of the time series \cite{ismail2020inceptiontime}. InceptionTime, on the other hand, is faster to train, but the stochastic nature of the process leads to high variability in the results, making it necessary to use an ensemble of networks for prediction \cite{ismail2020inceptiontime}.

To address these challenges, alternative models such as ROCKET, MiniRocket, and MultiRocket were introduced \cite{dempster2020rocket, dempster2021minirocket, tan2022multirocket}. They achieve state-of-the-art performance by training a shallow classifier on a large set of precomputed features generated by random convolutional kernels. Despite the significant reduction in learnable parameters, which greatly improves their training scalability, a large number of convolutions are still required to compute the features for inference. Considering the arbitrary nature of the kernels, we hypothesize that a significant fraction of the generated features are either redundant due to correlations or lack relevance for the classification task at hand. These superfluous features may not only add unnecessary computational load, but may also induce overfitting and compromise model generalization.

In this study, we present a novel methodology to sequentially select and prune the features generated by random convolutional kernel transformations in TSC, which we call Sequential Feature Detachment (SFD). Leveraging the linear nature of the classifier, our method uses model coefficients to estimate feature importance at each step. Unlike prior feature selection algorithms such as Thresholded Least-Squares (STLSQ) and Stepwise Sparse Regressor (SSR) \cite{brunton2016discovering, boninsegna2018sparse}, which were developed for sparse regression in the context of system identification, SFD is designed to efficiently handle tasks with large feature sets and eliminates the need for sensitive hyperparameter selection. Throughout this work, we will use the prefix 'Detach' to refer to SFD applied to any given base model (e.g., Detach-ROCKET).

To evaluate our methodology, we conducted experiments using the ROCKET model on the UCR archive, a standard benchmark for TSC. For the datasets studied, we show that models can be pruned to retain as little as $5\%$ of the original number of features while maintaining performance equal to or better than the full initial model. In particular, when retaining $10\%$ of the features, Detach-ROCKET shows on average a relative accuracy gain of $0.2\%$ compared to the full model. We also tested SFD in other variants of ROCKET, namely MiniRocket and MultiRocket, where it showed similar pruning performance. In a direct comparison with alternative pruning strategies for the ROCKET model, specifically S-ROCKET and POCKET, Detach-ROCKET achieved the same level of accuracy while utilizing only $10\%$ of the full model features, compared to an average of $35.10\%$ and $38.93\%$ for the other methods on the same datasets.


Additionally, we introduce an end-to-end procedure for finding the optimal model size when pruning a ROCKET model. This optimal configuration balances the trade-off between the number of pruned features and the accuracy of the pruned model on a validation set. Evaluation of the procedure on the largest binary UCR dataset reveals that it can reduce model size by $98.9\%$, while delivering an average increase in test accuracy of $0.6\%$ and attenuating overfitting in the training set from $99.8\%$ to $96.5\%$.

The rest of this paper is structured as follows. In Section 2, we review ROCKET models and previous work on feature selection. In Section 3, we introduce SFD and the end-to-end Detach-ROCKET model. In Section 4, we present our experimental results on the UCR dataset. In Section 5, we show a comparison of Detach-Rocket with alternative pruning strategies. In Section 6, we compute the complexity of our algorithm. Finally, Section 7 presents the discussion and Section 8 the conclusions.

\section{Background}
\label{sec:background}

\subsection{Time Series Classification}
Reflecting the dominant trend in machine learning, the majority of state-of-the-art solutions for TSC are characterized by models containing a large number of parameters \cite{ismail2019deep,bagnall2017great,ruiz2021great,middlehurst2023bake}. These include neural network architectures tailored for processing sequential data, such as RNNs, one-dimensional convolutions, and transformers \cite{yu2019review, uribarri2022dynamical, ismail2020inceptiontime, vaswani2017attention}. A particularly successful example is InceptionTime, a deep convolutional neural network inspired by Inception models for images \cite{ismail2020inceptiontime}. Another category with an extensive number of parameters comprises ensemble methods, which usually combine several different types of classifiers into a large meta-ensemble. flat-COTE, TS-CHIEF, HIVE-COTE 1.0, and HIVE-COTE 2.0 are prominent examples of this approach \cite{shifaz2020ts, bagnall2015time, lines2018time, middlehurst2021hive}.

Despite their high performance, large models are not without limitations, which become critical in sparse data scenarios or on systems constrained by computational resources. Their large dataset requirements, scalability challenges, and deployment complexity underscore the need for alternative approaches. Among the available alternatives, random convolution kernel models offer compelling performance.

\subsection{Random Convolutional Kernel Transformation (ROCKET)}
\label{sec:background_rocket}
ROCKET is a computationally efficient model designed for time series classification \cite{dempster2020rocket}. It employs a large set of $K$ non-trainable random convolutional kernels to process the input time series $\mathbf{s}_i$ and generate a collection of feature maps $\{\mathbf{f_{ik}}\}$ by applying the following convolution operation:

\begin{equation}
\label{eq:convolution}
\mathbf{f}_{ik} = \mathbf{s}_i * \mathbf{R}_k + b_k .
\end{equation}

The set of $K$ randomly generated kernels are characterized by their coefficient values $\{\mathbf{R_k}\}$, bias values $\{b_k\}$, and their associated convolution properties, such as dilation and padding. Then, from each feature map $\mathbf{f}_{ik}$, the maximum value (MAX) and the proportion of positive values (PPV) are extracted, yielding a total of $F=2K$ final attributes:

\begin{equation}
\label{eq:max}
\text{MAX}(\mathbf{f}_{ik}) = \max(\mathbf{f}_{ik}),
\end{equation}

\begin{equation}
\label{eq:ppv}
\text{PPV}(\mathbf{f}_{ik}) = \frac{\#\{ z \in \mathbf{f}_{ik} : z > 0 \}}{\#\{ z \in \mathbf{f}_{ik}\}},
\end{equation}

\begin{equation}
\label{eq:features}
\mathbf{x}_i = [\text{MAX}(\mathbf{f}_{i1}), \text{PPV}(\mathbf{f}_{i1}), \ldots, \text{MAX}(\mathbf{f}_{iK}), \text{PPV}(\mathbf{f}_{iK})].
\end{equation}

Here $\mathbf{x}_i$ refers to the feature set associated to the the input time series $\mathbf{s}_i$. After this transformation stage, which requires no training, a ridge linear classifier is fitted to the training dataset. In the binary case, the target labels $y_i$ are converted to $\{-1,1\}$ and the following optimization problem is solved:

\begin{equation}
\label{eq:ridge}
\hat{\theta}^{\mathrm{ridge}}=\underset{\theta}{\operatorname{argmin}}\left\{\sum_{i=1}^N\left(y_i-\theta_0-\sum_{k=1}^{2K} x_{i k} \theta_k\right)^2+\lambda \sum_{k=1}^{2K} \theta_k^2\right\},
\end{equation}

where $N$ is the size of the training dataset, and $\theta_k$ are the coefficients of the regression to be found. For smaller to medium-sized datasets, solving the ridge regression using matrix factorization techniques is usually the best option \cite{rifkin2007notes, dempster2020rocket}, while for larger datasets, an iterative method like stochastic gradient descent provides fast training with a fixed memory cost \cite{dempster2020rocket}.

The performance of ROCKET is comparable to other leading larger models, but it requires significantly less computation \cite{dempster2020rocket,ruiz2021great}. Since ROCKET's convolutional phase consists of non-learnable kernels, only the linear classifier needs to be trained. As a result, the training process is substantially simpler and faster, allowing ROCKET to handle larger datasets that were previously inaccessible to more complex models \cite{dempster2020rocket}. However, this efficiency may come at a cost. It is reasonable to expect that when using fixed (non-learnable) basis functions, the error in the worst-case scenario may be considerably larger compared to the use of adaptable (learnable) basis functions, as seen in neural networks. As shown in \cite{barron1993universal}, in a quite general setting, the lower bound for the integrated squared error of a linear combination of $n$ non-learnable basis functions scales as $\propto (1/n)^{(2/d)}$, where $d$ is the dimensionality of the space, which in our case is the length of the time series. Therefore, a large number of basis functions, or kernels in the context of ROCKET, is required for effective class discrimination.

Building on the foundation of ROCKET, numerous variants have emerged \cite{dempster2021minirocket, tan2022multirocket,schlegel2022hdc, wang2023ceemd, dempster2023hydra}, of which two are noteworthy: MiniRocket and MultiRocket. MiniRocket elaborates on the approach by employing a dictionary of fixed diverse kernels instead of randomly generating them, providing similar accuracy through a more efficient process. MultiRocket, on the other hand, advances the MiniRocket design by including the first order difference of the time series as part of the input and integrating a wider array of pooling operators for each feature map. This increases the expressiveness of the extracted features and improves model performance at the cost of greater computational load.

All variants of ROCKET, regardless of their differences, share the same core strategy. First, the time series is subjected to a non-learnable transformation stage that produces a massive amount of features. Then, a regularized linear classifier is trained within this high-dimensional representation space. While effective, this strategy is not optimal. In particular, due to the arbitrary generation process, numerous features may be irrelevant to the classification task. Furthermore, even among the relevant features, many could potentially be highly correlated due to their sheer number. This excess of redundant and non-informative features can not only create unnecessary computational overhead, but also increase the potential for overfitting.

\subsection{Previous Work}

In a previous study \cite{salehinejad2022s}, the authors investigated kernel selection and pruning for TSC with random convolutional kernels. They propose a population-based evolutionary optimization algorithm to find the optimal combination of kernels to keep in a ROCKET classifier, which they called S-ROCKET. By training a new ridge classifier on these selected features, they were able to achieve similar performance to the original full models while pruning $60\%$ of the kernels.

In contrast to the aforementioned work, which uses a model-agnostic approach to prune the less relevant features, our methodology exploits the simplicity and shallowness of the linear classifier employed in ROCKET models. The problem of feature selection within linear models has been extensively studied in the field of statistical learning, providing a broad body of knowledge to draw upon. Particularly, when the number of features suprasses the number of data samples, this problem is referred to as sparse regression or compressive sensing \cite{hastie2009elements, james2013introduction, tibshirani1996regression, donoho2006compressed, candes2006compressive}. The main challenge in our particular context lies in developing a methodology that can effectively handle the substantial number of features present in our models, with 20,000 in the case of the regular ROCKET model and up to 80,000 in MultiRocket. 

A standard solution to address feature selection in a linear model is to apply a lasso (L1) regularization to the linear classifier \cite{james2013introduction, tibshirani1996regression}. Then, by changing the regularization parameter, one could detect and discard the coefficients which are the ones first approaching zero as the regularization parameter increases. However, there is no closed-form expression for the solution of a lasso regression, and the available numerical algorithms for solving it are computationally expensive, making this option challenging for large datasets such as those produced by the ROCKET models. Nevertheless, in a recent preprint study, the authors explore the use of a particular variant of lasso regression, group elastic net, for feature selection in ROCKET models \cite{chen2023p}. They call their algorithm POCKET and show that it produces similar results to S-ROCKET.

An alternative strategy to Lasso regularization involves utilizing a backward-stepwise selection technique, which systematically removes less relevant features from the classification problem \cite{hastie2009elements}. Recent studies in system identification have showcased the effectiveness of two specific variations of this strategy, namely Sequential Thresholded Least-Squares (STLSQ) and Stepwise Sparse Regressor (SSR) \cite{brunton2016discovering, boninsegna2018sparse}. Both algorithms exhibit robust performance and usually outperform other alternatives such as lasso or forward-stepwise selection, especially in scenarios where datasets are large or noisy \cite{kaptanoglu2023benchmarking, de2020pysindy, kaptanoglu2021pysindy}. However, they are suboptimal for ROCKET feature selection because they require either a sensitive and unintuitive choice of hyperparameters or very heavy computation.

In the STLSQ algorithm, the one used in the popular SINDy method for systems identification, features are sequentially eliminated at each step if their corresponding regression coefficient does not exceed a predetermined threshold value. The elimination process ends when all remaining coefficients exceed the threshold \cite{brunton2016discovering}. This stands in contrast to SSR, where the feature with the smallest coefficient is discarded at each step until a model with only one feature remains. Subsequently, STLSQ applies a cross-validation procedure to select the optimal step and estimate the most suitable model configuration \cite{boninsegna2018sparse}.

In the following section, we present our proposed feature selection algorithm called Sequential Feature Detachment (SFD), which is a novel variant of the backward-stepwise selection technique specifically designed to address the computational challenge of the large number of features present in the classification problem after a ROCKET transformation.

\section{Methods}
\label{sec:methods}

In this section, we provide a comprehensive description of the methods used in this paper. Section \ref{sec:sfd} introduces SFD, the proposed algorithm for performing feature selection in linear classifiers with a large number of features. Section \ref{sec:end2end} describes the end-to-end procedure for training a pruned ROCKET model, including the criteria for selecting the optimal model size.

\subsection{Sequential Feature Detachment (SFD)}
\label{sec:sfd}

Let $\boldsymbol{\mathcal{F}}$ denote the set of features produced by the ROCKET transformation, which has a cardinality $F=2K=\# \boldsymbol{\mathcal{F}}$. In the case of a standard ROCKET model, $F=20000$, while for MultiRocket $F=80000$. The dataset to be used must be normalized along these features. We define $\boldsymbol{\mathcal{S}} \subseteq \boldsymbol{\mathcal{F}}$ as the set of active features, which is initialized to be equal to $\boldsymbol{\mathcal{F}}$ and updated at each step.

The feature selection procedure involves iterative detachment steps, where a fixed percentage of the current active features is removed from $\boldsymbol{\mathcal{S}}$ at each step. The proportion of removed features, denoted as $p$, is the only relevant hyperparameter in the process. 

At each step $t$, a ridge classifier is trained on the set of active features at that step $\boldsymbol{\mathcal{S}}_{t}$. This is done by solving the following optimization problem:

\begin{equation}
\label{eq:ridge_sfd}
\hat{\theta}_t^{\mathrm{ridge}}=\underset{\theta}{\operatorname{argmin}}\left\{\sum_{i=1}^N\left(y_i-\theta_0-\sum_{k \in \boldsymbol{\mathcal{S}}_t} x_{i k} \theta_k\right)^2+\lambda \sum_{k \in \boldsymbol{\mathcal{S}}_t} \theta_k^2\right\}
\end{equation}

The training yields a set of optimal coefficients, $\hat{\theta}_t^{\mathrm{ridge}} = \{\hat{\theta}_k$\}, each of which is associated with one particular active feature. The features are subsequently ranked based on the absolute value of these coefficients, \{$|\hat{\theta}_k|$\}. These values are proportional to the influence of each feature on the classifier's decision. The lowest $100\cdot p$ percentage of ranked features are then discarded, while the remaining $100\cdot (1-p)$ percentage is retained as the new set of active features $\boldsymbol{\mathcal{S}}_{t+1}$. The pseudocode for this process is presented in Algorithm \ref{alg:sfd}. Note that during this detachment process, the regularization parameter $\lambda$ of the ridge regressors is set to the value obtained for the initial full model using cross-validation.

\begin{algorithm}
  \caption{Sequential Feature Detachment}
  \label{alg:sfd}
  \begin{algorithmic}
	\STATE \textbf{Parameters:}
  	\STATE $M$: Number of steps
  	\STATE $N$: Number of initial features
        \STATE $K$: Number of kernels
  	\STATE $p$: Proportion of eliminated features at each step
	\STATE
	\STATE \textbf{initialize:} ROCKET model with $K$ kernels
	\STATE \textbf{initialize:} Active feature set $\boldsymbol{\mathcal{S}}$ with $F=2K$ features
	\STATE Train ridge classifier with LOOCV to find $\lambda$ value
	\FOR{$t = 1$ to $M$}
        \STATE Train ridge classifier on $\boldsymbol{\mathcal{S}}$ and obtain optimal coefficient ${\hat{\theta}_k}$ for each active feature
  	\STATE Rank features based on $|\hat{\theta}_k|$
  	\STATE Discard lowest $p$ fraction of ranked features
  	\STATE Update active feature set $\boldsymbol{\mathcal{S}}$ with retained features  
	\ENDFOR
	\RETURN{} Selected features at each step $\boldsymbol{\mathcal{S}}_t$
	\end{algorithmic}
\end{algorithm}

The value of the parameter $p$ controls the trade-off between the computational cost and precision in the procedure. A higher $p$ value speeds up pruning, but increases the risk of eliminating an entire family of correlated features that encode similar but relevant information from the time series. Note that in the case of correlated features, the magnitude of their individual coefficients may be small because the weight of their contribution to the classifier is distributed among all of them. To maintain a conservative default value, we set $p=0.05$, which corresponds to removing $5\%$ of the features at each step. In scenarios where computational training time or procedure precision is a concern, the value can be adjusted.

The percentage of total features removed at step $t$ is independent of the number of initial features $F$, and it is given by the expression $1-(1-p)^{t}$. As the number $t$ increases, the computational cost of each step decreases as the trained model becomes smaller. The total number of steps performed, $M$, determines the maximum proportion of pruned features explored during the selection process. The choice of $M$ is not critical, as long as the number of removed features extends beyond the optimal. For $p=0.05$, setting $M=150$ will explore removals of up to $99.95\%$ of the features, which should be sufficient for most applications.

This methodology can also be extended to the case of multi-class problems, where there are more than 2 possible classes for each instance. Given a classification problem with $C$ number of classes, $C$ ridge classifiers are trained simultaneously, and each feature $k$ now has $C$ associated coefficients, $\hat{\theta}_{kc}$. To determine the relevance of each feature and rank them accordingly, we implement a max-pooling operation along the class dimension. That is, at each step $t$, the features are ranked based on the values \{$|{\operatorname{max}}_{c}(\hat{\theta}_{kc})|$\} instead of \{$|\hat{\theta}_k|$\}, and the rest of the procedure remains the same as for the binary case.

\subsection{End-to-End Feature Pruning Optimization}
\label{sec:end2end}

To determine the optimal number of features to retain in a given dataset, we introduce an end-to-end procedure. The procedure starts by dividing the dataset into three subsets: a training set, a validation set, and a test set. A standard ROCKET model is then trained on the training set and the SFD process is applied (as described in section \ref{sec:sfd}). Evaluation of the model at each pruning step on the validation set yields an accuracy curve. To select the optimal fraction of pruned features on this curve, $Q_c$, a straightforward optimization problem is solved:

\begin{equation}
\label{eq:tradeoff}
\begin{split}
Q_c &= \operatorname*{argmax}_{q} \ f_c(q) = \operatorname*{argmax}_{q} \left\{ \alpha(q)+c \cdot q \right\}.
\end{split}
\end{equation}

In this equation, $q$ represents the proportion of pruned features and $\alpha(q)$ the accuracy of the pruned model on the validation set. The hyperparameter $c$ serves as a weighting factor that determines the tradeoff between these two variables which we want to maximize. The choice of $c$ depends on the specific requirements of the use-case, with smaller values favoring accuracy and larger values favoring size.

Once $Q_c$ is determined, the selected features in the pruned ROCKET are used to retrain a ridge classifier on the combined training and validation sets. This results in the final Detach-ROCKET model, an optimal reduced model containing only a subset of the original kernels. A final evaluation of Detach-ROCKET is then performed on the test dataset.

\section{Experiments}
\label{sec:experiments}

\subsection{Datasets and Model Description}

To test the SFD methodology, we used all 42 binary classification datasets available in the UCR Time Series Classification Archive \cite{dau2019ucr}, a widely recognized benchmark for time series classification \cite{bagnall2017great, middlehurst2023bake}. These datasets cover a wide range of univariate time series, varying in size and characteristics. Our decision to focus exclusively on binary classification tasks was primarily driven by computational constraints and a desire to evaluate SFD in its simplest setting, without class mixing. It is important to note that, as described in Section \ref{sec:sfd}, the SFD algorithm can also handle multiclass scenarios. In Section \ref{sec:alt_methods}, we compare Detach-ROCKET to alternative methods in a benchmark that includes both binary and multiclass problems.

In the feature generation phase of our experiments, we employed the standard ROCKET algorithm with a kernel count of $K=10000$, as recommended by its original developers. For the implementation we used the sktime library \cite{loning2019sktime, markus_loning_2022_7117735}. This transformation resulted in each input time series being mapped to a 20000-dimensional feature vector, representing the PPV and MAX values of the convolution with each kernel. Subsequently, a ridge classifier was trained on these generated features. For training, we respected the original training/test split for each dataset. The regularization parameter for this classifier was determined using Leave One Out cross-validation across 20 alpha values, uniformly distributed on a logarithmic scale between $e^{-10}$ and $e^{10}$. In the text, we refer to this classifier as the Full Model.

All experiments were performed with the default SFD parameters, comprising the proportion of removed features $p$ set to $0.05$ and number of iteration steps $M$ set to $150$.

\subsection{SFD Results}

We evaluated the SFD process on the 42 binary datasets from the UCR archive. The feature detachment process was applied 25 times to each dataset, using a different starting set of random kernels for each run. To evaluate its performance, we compare the accuracy obtained on the test set by the pruned model with the accuracy obtained on the test set by the full ROCKET model.

\subsubsection{Comparative Performance of Pruned Models to Full ROCKET}

\begin{figure}[ht]
  \centering
  \includegraphics[width=0.7\textwidth]{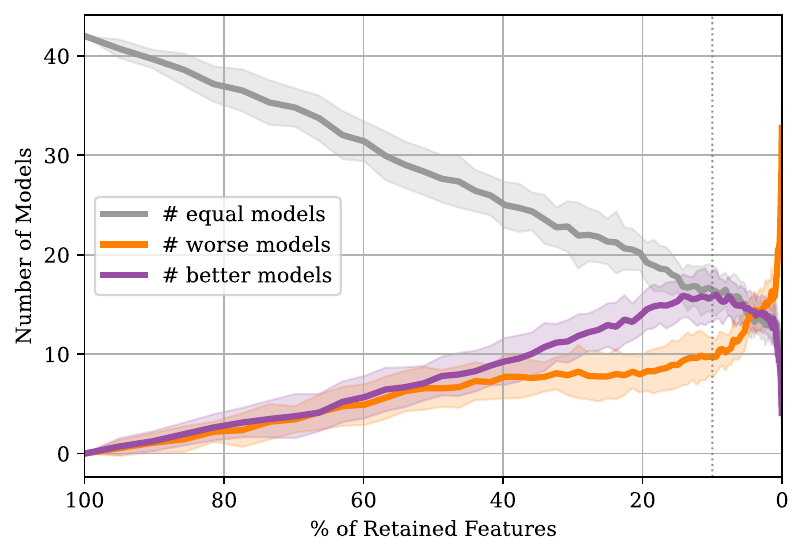}
  \caption{Performance comparison of pruned models using SFD relative to the full ROCKET model. The plot shows the number of pruned models with better, equal, and worse performance compared to the full model, plotted against the percentage of retained features. The solid lines represent the mean number over 25 iterations, while the shaded area indicates the standard deviation. The dashed vertical line marks $10\%$ of retained features.}
  \label{fig:fig1}
\end{figure}

Figure \ref{fig:fig1} shows the number of pruned models with better, equal, and worse performance compared to the full model, plotted against the percentage of retained features. The mean across the 25 iterations is depicted by solid lines, with the shaded region denoting the standard deviation. At the beginning of the feature detachment process, all pruned models (across the 42 datasets) mirror the performance of the full ROCKET model. As the feature reduction progresses, there is an even division among the models that become better or worse compared to the full model. Once the percentage of retained features drops below $40\%$, the pruned models that outperform the full model become more prevalent. Specifically, when feature retention is between $35.85\%$ and $5.10\%$, there are significantly more pruned models that outperform the full model than pruned models that perform worse than the full model, as confirmed by a Kolmogorov–Smirnov (KS) test with $p<0.05$. Results of the KS test are presented in Figure \ref{fig:fig5} of the Appendix section. 

Conversely, after $95\%$ feature reduction, there is a noticeable drop in the number of pruned models that outperform the full model. With less than $1.73\%$ of features remaining, the underperforming models significantly outnumber the better-performing models (KS test with $p<0.05$, see Figure \ref{fig:fig5}). This drop in performance is to be expected when only a few features remain, since many classification tasks will not be linearly separable in a highly reduced feature space.

It is noteworthy that many pruned models, despite extensive feature elimination, maintain the exact same accuracy as the original full ROCKET model. This is due to two primary reasons: First, the full ROCKET model achieves $100\%$ test set accuracy in 5 of the 42 datasets, making it impossible for the pruned model to perform better. Second, certain datasets have a limited number of test set instances, resulting in insensitive accuracy metrics and thus identical results.

\subsubsection{Quantification of Accuracy Change Relative to Feature Retention}

\begin{figure}[ht]
  \centering
  \includegraphics[width=0.7\textwidth]{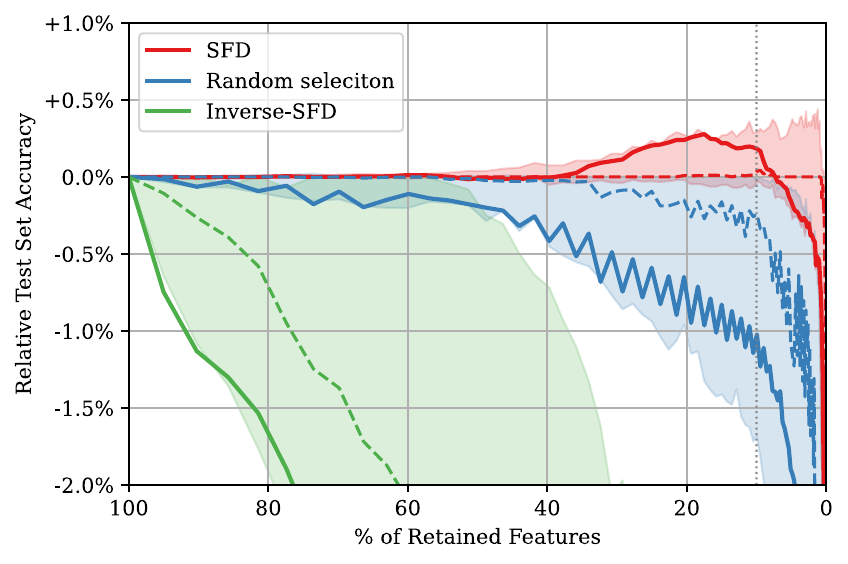}
  \caption{Percentage change in test set accuracy (relative to the full ROCKET model) of pruned models using SFD as a function of retained features. Results are shown for the 42 studied datasets, each averaged over 25 realizations (1050 realizations in total). The solid line represents the mean, the dashed line represents the median, and the shaded area represents the interquartile range (25th to 75th percentile). For comparison, the figure also includes two alternative pruning strategies: random selection (in blue) and inverse-SFD (in green).}
  \label{fig:fig2}
\end{figure}

To quantify the percentage gained or lost by the pruned models compared to the full model, we computed the percentage change in accuracy on the test set as a function of the features retained. Figure \ref{fig:fig2} displays the results for the 42 datasets, with each represented as the mean over 25 realizations (1050 realizations in total).  We provide statistics including the mean, median, and quartiles (25th and 75th percentiles) across all 42 datasets. The graph reveals that between a level of $30\%$ and $10\%$ of feature retention, the gains from the improved pruned models outweigh the losses from the degraded models. Since many of the pruned models match the performance of the full models, the median of the percentage change is zero for most of the pruning process. Given that the average accuracy of the full ROCKET models across all realizations is $91\%$, many of the pruned models have limited potential for improvement. This limitation is reflected in the skewed distribution of percentage change and the modest percentage gains over the full model.

Another interesting aspect to evaluate is the relevance of the optimal selection of features to discard during the pruning process. For this purpose, we present two alternative pruning strategies: one that randomly removes $5\%$ of the features at each stage (random selection) and another that removes the top $5\%$ most informative features (inverse-SFD). The comparison of SFD with both alternatives is shown in Figure \ref{fig:fig2}. On average, SFD outperforms random selection, while inverse-SFD is clearly the worst of the three. This result shows that SFD captures a hierarchy of importance in the features, and further supports the hypothesis that only a subset of features produced by ROCKET are relevant to the classification task.

Some datasets maintain strong performance even when a considerable percentage of features are randomly removed, suggesting that some binary classification tasks in UCR are relatively straightforward and can be efficiently handled by ROCKET models with a smaller number of kernels.

\subsection{Detach-ROCKET with 10\% of Retained Features}

\begin{figure}[htb]
  \centering
  \includegraphics[width=1.0\textwidth]{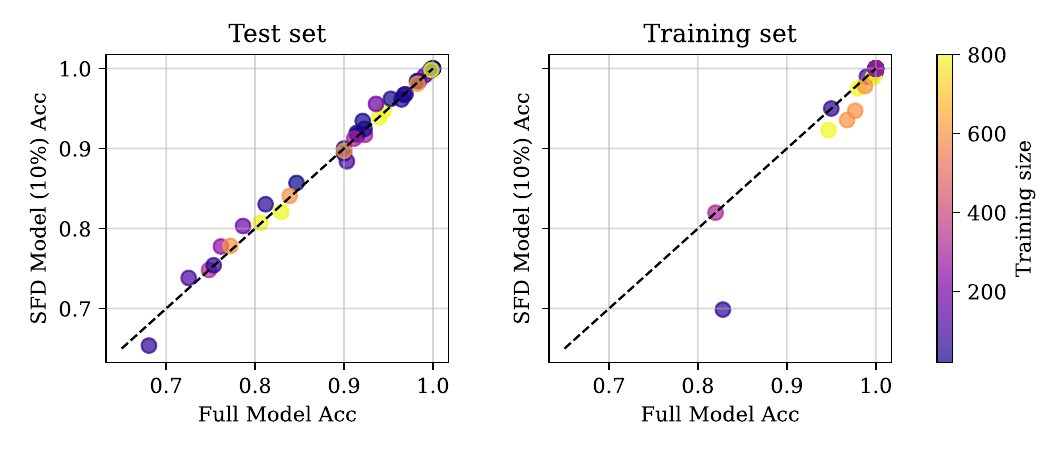}
  \caption{Accuracy (Acc) comparison between the pruned models with $10\%$ of retained features and the full ROCKET model, for both test and training sets. Each point depicts the average accuracy of one of the 42 datasets over 25 realizations. The color indicates the training size of the dataset, with the color range saturated at 800 instances for enhanced contrast.}
  \label{fig:fig3}
\end{figure}

When employing SFD to prune a ROCKET classifier, the optimal final percentage of features to retain will ultimately depend on the dataset and the intended use case. In the following section, we show the results of our proposed method to determine this optimal number of features using a validation set. This selection is governed by a hyperparameter that controls the tradeoff between accuracy and model size. 

Considering the results obtained for the binary UCR datasets, when a validation split is not feasible, we recommend retaining $10\%$ of features for a standard ROCKET with $10,000$ kernels ($20,000$ features). This recommendation is based on the results presented in Figure \ref{fig:fig1}, where it is shown that the proportion of pruned models that are better than the full models peaks around $10\%$.

Figure \ref{fig:fig3} shows the accuracy comparison between the $10\%$ pruned model and the full ROCKET model, in both test and training sets. Each point represents the average accuracy of a data set over 25 realizations. The pruned models performed better for $52\%$ of the datasets, equally well for $17\%$, and worse for $31\%$, with an average accuracy improvement of $0.2\%$. Although this selection is suboptimal, it can be useful in scenarios where the number of training samples is too small to separate a representative validation set. Another notable aspect of the pruned models is their reduction in accuracy on the training set, indicating a reduction in overfitting compared to the full model. The detailed results for each of the datasets can be found in Table \ref{tab:detailed_results} in the Appendix section.

In addition, we tested the effectiveness of our prunning methodology on other relevant variants of random convolutional kernel models: MiniRocket and MultiRocket. The results comparing the accuracies of the full models and their corresponding pruned versions are presented in Table \ref{tab:detailed_results} of the Appendix section. 

In the case of Detach-MiniRocket, the average relative change in accuracy when retaining $10\%$ of the features was $-0.4\%$. In contrast to the ROCKET case, there is a slight decrease in accuracy for the pruned models. This can be explained by the fact that MiniRocket starts with an initial set of features that is smaller ($10,000$) and less redundant, and after pruning $90\%$ of them, we keep half the amount of features compared to ROCKET.

The default MultiRocket transformation produces a substantially larger initial feature pool than ROCKET ($80,000$). To compensate for this difference, we set the fixed percentage of retained features in our experiments to be $5\%$. The resulting average relative change in accuracy for Detach-MultiRocket was $0.4\%$, twice that of Detach-ROCKET. In contrast to the MiniRocket case, the larger set of features created with MultiRocket seems to benefit more from feature selection and pruning.

\subsection{Detach-ROCKET end-to-end Procedure}
\label{sec:end-to-end}

\begin{figure}[ht]
  \centering
  \includegraphics[width=0.7\textwidth]{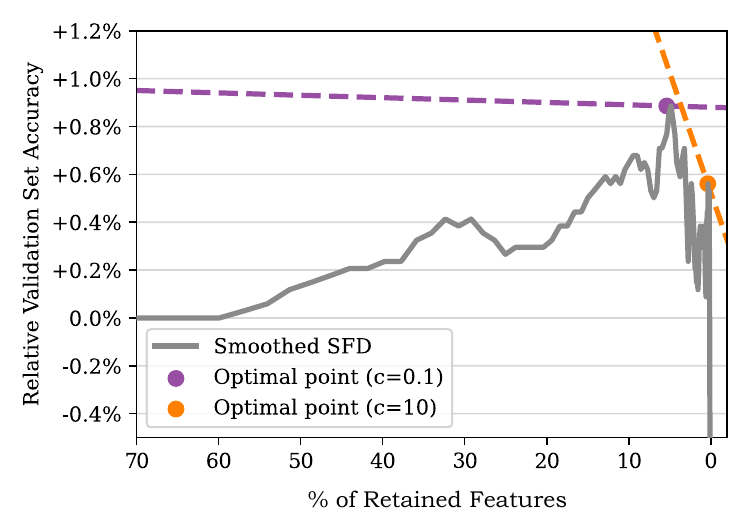}
  \caption{Selection of optimal number of features using Detach-ROCKET. The grey curve represents the validation set accuracy achieved by the ridge classifier as a function of the percentage of retained features. Optimal values corresponding to $c=0.1$ and $c=10$ are highlighted in violet and orange, respectively. Dashed lines indicate the level sets for the optimized objective functions (see Equation \ref{eq:tradeoff}), showing that the function value changes perpendicularly to these lines.}
  \label{fig:fig4}
\end{figure}

To conduct a comprehensive evaluation of the end-to-end procedure, we selected the largest dataset within the UCR binary set, \emph{FordB}, ensuring an adequate number of samples in each of the training, validation, and test sets. The goal in \emph{FordB} is to diagnose whether or not a particular symptom is present in an automotive subsystem by analyzing engine noise. As provided by UCR, the dataset comprises a training set with $3636$ samples and a test set with $810$ samples. For validation, we employed a random subsampling method to select $33\%$ of the elements from the training set.

Figure \ref{fig:fig4} illustrates the feature count optimization for one of the pruned models, where we show the accuracy on the validation set as a function of the percentage of retained features and the selected optimal points for weighting factors $c=0.1$ and $c=10$.

To control for stochastic variability arising from training set subsampling and kernel generation, the experiment to evaluate the end-to-end procedure was conducted 25 times. Results are summarized in Table \ref{tab:detach_rocket}. For Detach-ROCKET with $c=0.1$, all 25 pruned models consistently surpassed the accuracy of their corresponding full ROCKET models. The mean test set accuracy shows an improvement of $0.68\%$, validated by a KS test with $p=0.01$. For Detach-ROCKET models with $c=1$ and $c=10$, the average accuracy differed by $0.41\%$ and $-0.5\%$, respectively. In these latter scenarios, however, the difference relative to the full ROCKET model was not statistically significant ($ \text{KS test } p>0.05$). In Table \ref{tab:detach_rocket} we also present the performance of Small ROCKET models. These are regular ROCKET models with a number of features equal to the average number of features of the Detach-ROCKET models. These models perform significantly worse than the Detach-ROCKET models, highlighting the benefits of starting with a large model and reducing it with SFD, rather than directly training a small ROCKET model from scratch.

\begin{table}[h]
  \centering
  \begin{tabular}{llll}
	\toprule
	Model 	& Test accuracy (\%) [$\uparrow$]	& Train accuracy (\%) & Number of Features [$\downarrow$] \\
	\midrule\midrule

	ROCKET (K=10000) & $80.37 \pm 0.61$ & $99.77 \pm 0.04$  & $20000 \ \  (100\%)$ \\
 \midrule
	Detach-ROCKET (c=0.1) 	& $\mathbf{81.05 \pm 0.64}$ & $96.51 \pm 1.04$  & $228.66 \pm 174.64 \ \  (1.14\% \pm 0.87\%)$\\
	Detach-ROCKET (c=1) 	& $80.78 \pm 0.79$ & $95.22 \pm 0.49$  & $52.29 \pm 26.94 \ \  (0.26\% \pm 0.13\%)$ \\
	Detach-ROCKET (c=10) 	& $79.87 \pm 1.22$ & $94.27 \pm 0.32$  & $\mathbf{16.24 \pm 4.10 \ \  (0.08\% \pm 0.02\%)}$ \\
 \midrule
 	Small ROCKET (K=114) 	& $77.61 \pm 1.00$ & $94.16 \pm 0.44$  & $228 \ \  (1.14\%)$ \\
 	Small ROCKET (K=26) 	& $72.81 \pm 2.38$ & $89.38 \pm 1.34$  & $52 \ \  (0.26\%)$ \\
 	Small ROCKET (K=8) 	& $64.65 \pm 4.21$ & $80.18 \pm 4.13$  & $16 \ \  (0.08\%)$ \\
	\bottomrule
  \end{tabular}
  \vspace*{2mm}
   \caption{Summary of classification accuracy across 25 repetitions for ROCKET and Detach-ROCKET models with $c=0.1, 1, 10$ on the FordB dataset from the UCR archive. Performance for Small ROCKET models with the same number of features as the Detach-ROCKET models is also presented.  Results are reported as mean $\pm$ standard deviation.}
  \label{tab:detach_rocket}
\end{table}

\section{Comparison with Alternative Methodologies}
\label{sec:alt_methods}
We evaluated the performance of Detach-ROCKET against state-of-the-art pruning methods, specifically Small ROCKET or S-ROCKET \cite{salehinejad2022s} and Pruned ROCKET or POCKET \cite{chen2023p}. To ensure a fair comparison, we utilized the first 30 datasets from the UCR catalog, consistent with the benchmarks performed in their respective works. 

The results, formatted as an extension of those presented in the POCKET paper, are shown in Table \ref{tab:comparison}. The pruning rate for Detach-ROCKET was set to 10\%, as determined by the analysis in the previous section. Additionally, we included results for the end-to-end version of Detach ROCKET, where the pruning rate is set automatically.

In terms of accuracy, all methods demonstrate similar performance when considering deviations. However, they operate at different levels of retained features. On average, S-ROCKET retains 39\% of the features. POCKET does not have an automated method for setting the retention rate and instead uses the values determined by S-ROCKET, with exceptions for the first and last records, resulting in an average retention rate of 35\%. In contrast, our methodology maintains the same accuracy with a fixed pruning level of 10\% and can be pushed even further to retain only 2\% of the features in the end-to-end mode.

It is worth noting that the end-to-end methodology relies on having a representative validation set to select the optimal model size. This condition is not always met by UCR datasets, where some datasets have a small training size. Nevertheless, we have included these results to show that even under these suboptimal settings, Detach-ROCKET is able to prune up to $98\%$ of the total features without significant loss of accuracy.

\begin{table*}[ht]
    \centering
    \setlength{\tabcolsep}{2pt}
    \renewcommand{\arraystretch}{1.2}
    \begin{adjustbox}{width={\textwidth},totalheight={\textheight},keepaspectratio}%
 \begin{tabular}{l|c|cc|cc|cc|cc}
    \hline

    & \multicolumn{1}{|c|}{Original ROCKET [\cite{dempster2020rocket}]} & \multicolumn{2}{|c|}{S-ROCKET [\cite{salehinejad2022s}]} & \multicolumn{2}{|c|}{POCKET [\cite{chen2023p}]} & \multicolumn{2}{|c|}{\textbf{D-ROCKET} (fixed 10\%)} & \multicolumn{2}{|c}{\textbf{D-ROCKET} (c=1)}\\
    
    
    
    Dataset (Train Size)& Test Acc.(\%)& Features (\%) & Test Acc.(\%)& Features (\%) & Test Acc.(\%)& Features (\%)& Test Acc.(\%)& Features (\%)& Test Acc.(\%)\\
    
    \midrule\midrule
    
    Adiac (390)& 78.13$\pm$0.43 & 100.00 & 78.13$\pm$0.43 & 50.00 & 79.87$\pm$0.50 & 10 & 79.64$\pm$0.85 & 2.78 & 77.47$\pm$1.30 \\
    ArrowHead (36)& 81.37$\pm$1.03 & 24.47 & 81.77$\pm$1.31 & 24.47 & 81.83$\pm$1.59 & 10 & 82.91$\pm$1.29 & 0.46 & 82.17$\pm$2.55 \\
    Beef (30)& 82.00$\pm$3.71 & 19.46 & 81.00$\pm$2.60 & 19.46 & 83.33$\pm$3.65 & 10 & 83.00$\pm$1.00 & 0.35 & 76.00$\pm$6.63 \\
    BeetleFly (20)& 90.00$\pm$0.00 & 21.08 & 90.00$\pm$0.00 & 21.08 & 90.00$\pm$0.00 & 10 & 89.50$\pm$1.50 & 0.44 & 80.50$\pm$9.86 \\
    BirdChicken (20)& 90.00$\pm$0.00 & 24.31 & 89.00$\pm$3.00 & 24.31 & 90.00$\pm$0.00 & 10 & 90.00$\pm$0.00 & 0.10 & 88.50$\pm$5.02 \\
    Car (60)& 88.33$\pm$1.83 & 34.29 & 88.33$\pm$2.69 & 34.29 & 91.67$\pm$1.29 & 10 & 92.67$\pm$0.82 & 1.04 & 87.50$\pm$4.79 \\
    CBF (30)& 100.00$\pm$0.00 & 17.98 & 99.96$\pm$0.05 & 17.98 & 99.98$\pm$0.04 & 10 & 99.91$\pm$0.04 & 0.06 & 97.84$\pm$1.50 \\
    CinCECGT (40)& 83.61$\pm$0.55 & 24.44 & 82.79$\pm$0.74 & 24.44 & 88.23$\pm$1.64 & 10 & 90.71$\pm$0.52 & 0.18 & 91.04$\pm$2.71 \\
    ChlCon (467)& 81.50$\pm$0.49 & 40.67 & 79.26$\pm$1.46 & 40.67 & 80.71$\pm$0.60 & 10 & 78.21$\pm$0.62 & 3.92 & 73.00$\pm$2.80 \\
    Coffee (28)& 100.00$\pm$0.00 & 18.06 & 100.00$\pm$0.00 & 18.06 & 100.00$\pm$0.00 & 10 & 100.00$\pm$0.00 & 0.06 & 99.29$\pm$1.43 \\
    Computers (250)& 76.32$\pm$0.84 & 26.75 & 76.80$\pm$0.95 & 26.75 & 77.20$\pm$0.76 & 10 & 75.04$\pm$1.05 & 2.49 & 77.32$\pm$1.66 \\
    CricketX (390)& 81.92$\pm$0.49 & 72.95 & 82.05$\pm$0.62 & 72.95 & 82.18$\pm$0.80 & 10 & 82.26$\pm$1.32 & 2.67 & 77.67$\pm$2.75 \\
    CricketY (390)& 85.38$\pm$0.60 & 57.50 & 85.08$\pm$0.56 & 52.50 & 84.90$\pm$0.71 & 10 & 83.26$\pm$0.84 & 2.42 & 79.72$\pm$3.85 \\
    CricketZ (390)& 85.44$\pm$0.64 & 70.40 & 85.03$\pm$0.69 & 70.40 & 85.10$\pm$0.71 & 10 & 83.41$\pm$0.81 & 3.05 & 81.10$\pm$1.00 \\
    DiaSizRed (16)& 97.09$\pm$0.61 & 24.23 & 96.50$\pm$0.84 & 24.23 & 97.84$\pm$0.55 & 10 & 97.19$\pm$0.73 & 10.0 (*)& 97.19$\pm$0.73 \\
    DisPhaOAG (400)& 75.68$\pm$0.63 & 32.85 & 74.89$\pm$0.99 & 32.85 & 73.74$\pm$0.98 & 10 & 73.38$\pm$2.18 & 1.77 & 74.68$\pm$1.90 \\
    DisPhaOutCor (600)& 76.74$\pm$0.88 & 32.71 & 77.03$\pm$1.30 & 32.71 & 76.27$\pm$1.61 & 10 & 78.04$\pm$1.43 & 1.49 & 76.88$\pm$0.53 \\
    DoLoDay (78)& 60.50$\pm$1.70 & 55.20 & 60.25$\pm$1.84 & 50.20 & 60.62$\pm$2.11 & 10 & 60.52$\pm$2.26 & 1.25 & 56.23$\pm$4.27 \\
    DoLoGam (20)& 86.09$\pm$0.54 & 18.43 & 86.45$\pm$0.80 & 18.43 & 88.26$\pm$0.63 & 10 & 86.06$\pm$0.87 & 2.41 & 78.58$\pm$8.21 \\
    DoLoWKE (20)& 97.68$\pm$0.29 & 18.28 & 97.54$\pm$0.35 & 18.28 & 98.48$\pm$0.22 & 10 & 98.41$\pm$0.00 & 0.06 & 93.89$\pm$10.00 \\
    Earthquakes (322)& 74.82$\pm$0.00 & 32.64 & 74.82$\pm$0.00 & 32.64 & 74.96$\pm$0.54 & 10 & 74.82$\pm$0.00 & 0.05 & 74.82$\pm$0.00 \\
    ECG200 (100)& 90.40$\pm$0.49 & 10.88 & 89.90$\pm$0.70 & 10.88 & 90.60$\pm$0.66 & 10 & 90.60$\pm$0.80 & 3.41 & 91.30$\pm$0.64 \\
    ECG5000 (500)& 94.75$\pm$0.05 & 33.91 & 94.68$\pm$0.09 & 33.91 & 94.78$\pm$0.06 & 10 & 94.74$\pm$0.17 & 0.07 & 92.97$\pm$0.26 \\
    ECGFiveDays (23)& 100.00$\pm$0.00 & 22.19 & 100.00$\pm$0.00 & 22.19 & 100.00$\pm$0.00 & 10 & 100.00$\pm$0.00 & 0.05 & 99.87$\pm$0.27 \\
    EOGHSignal (362)& 58.26$\pm$1.05 & 69.33 & 58.01$\pm$1.19 & 64.33 & 58.45$\pm$1.18 & 10 & 61.74$\pm$1.30 & 1.62 & 59.23$\pm$1.59 \\
    EOGVSignal (362)& 54.70$\pm$0.58 & 51.56 & 55.03$\pm$0.91 & 51.56 & 54.81$\pm$0.58 & 10 & 55.47$\pm$1.39 & 3.36 & 54.53$\pm$1.02 \\
    FaceAll (500)& 94.68$\pm$0.40 & 46.61 & 94.64$\pm$0.32 & 46.61 & 94.14$\pm$0.78 & 10 & 94.25$\pm$0.53 & 3.07 & 94.11$\pm$0.55 \\
    FaceFour (24)& 97.61$\pm$0.34 & 18.30 & 97.61$\pm$0.34 & 18.30 & 97.84$\pm$0.34 & 10 & 99.89$\pm$0.34 & 1.52 & 95.80$\pm$3.45 \\
    FacesUCR (200)& 96.20$\pm$0.09 & 48.43 & 96.20$\pm$0.08 & 48.43 & 96.34$\pm$0.08 & 10 & 96.48$\pm$0.15 & 1.69 & 95.23$\pm$0.51 \\
    FiftyWords (450)& 82.99$\pm$0.41 & 100.00 & 82.99$\pm$0.41 & 50.00 & 82.46$\pm$0.47 & 10 & 82.51$\pm$0.55 & 10.0 (*)& 82.51$\pm$0.55 \\
    
    \midrule\midrule
    \textbf{AVERAGE} & 84.74 \ \ \ 0.62 & 38.93 & 84.52 \ \ \ 0.84 & 35.10& 85.15 \ \ \ 0.77 & \textbf{10} & \textbf{85.15 \ \ \ 0.78} & 2.06 & 82.90 \ \ \ 2.74 \\
    \hline
\end{tabular}
\end{adjustbox}
    \caption{Comparison of classification accuracy and retention rate between Original ROCKET, S-ROCKET, POCKET, Detach-ROCKET with a fixed retention rate of 10\%, and End-to-End Detach-ROCKET. Accuracies are reported as mean $\pm$ standard deviation. (*) It is not possible to create a stratified validation set for these datasets because there is at least one class with only one instance in the training set, so the results for using a fixed 10\% are presented instead.}
    \label{tab:comparison}
    
\end{table*}

\section{Complexity, Training Time and Code Availability}
\label{sec:complexity}

\subsection{Complexity Analysis}

We introduced the Selective Feature Detachment as a novel methodology for feature selection. To understand the computational implications of employing this methodology, we need to break down the complexity of the associated algorithms into its subcomponents. Let the following variables be defined:

\begin{itemize}
    \item \( N \): Number of training instances
    \item \( K \): Number of kernels
    \item \( F \): Number of features
    \item \( L \): Length of the time series
    \item \( M \): Number of steps, which suggested default value is 150
\end{itemize}

According to the original article \cite{dempster2020rocket}, the ROCKET transformation stage, which is required as an initial step, has a complexity of:
\begin{equation}
O_{ROCKET} = O\left( K \cdot N \cdot L \right).
\end{equation}

In our implementation, which is the one provided in the scikit-learn library, the complexity of the ridge classifier is governed by the relationship between the number of features $F$ and the number of training instances $N$ \cite{scikit-learn,dempster2020rocket,dongarra2018singular}. Specifically, the solving method alternates between eigen decomposition and singular value decomposition, with an effective complexity of:

\begin{equation}
O_{Ridge} = 
\begin{cases} 
O( N \cdot F^2 ), & \text{when } F < N \\
O( N^2 \cdot F ), & \text{when } F > N 
\end{cases}
\end{equation}

The overall complexity of SFD is determined by the sum of the complexities over the $M$ iteration steps. At each step $t$, the dominant computation is the training of the ridge classifier. Let $F_t$ denote the number of features retained at step $t$, then the complexity of SFD can be expressed as:

\begin{equation}
O_{SFD} =  \sum_{t=1}^{M} O\left( \min\{N,F_t\}^2 \cdot \max\{N,F_t\} \right).
\end{equation}

Considering that $F=F_{0}$ and $ F_{t+1} < F_t $, then $F$ is the largest value in $F_t$, and an upper bound for the complexity of the algorithm is given by:

\begin{equation}
O_{SFD} = 
\begin{cases} 
\sum_{j=1}^{M} O(N^2 \cdot F) = M \cdot O(N^2 F) = O(M \cdot N^2 \cdot F), & \text{when } N \leq F \\
\sum_{j=1}^{M} O(F^2 \cdot N) = M \cdot O(F^2 N) = O(M \cdot F^2 \cdot N), & \text{when } F \leq N
\end{cases}
\end{equation}

\subsection{Training Time}

In our end-to-end experiments presented in section $\ref{sec:end-to-end}$, we evaluated the computational overhead introduced by the SFD pruning methodology. While the training time for the full ROCKET was 305 seconds on average, the inclusion of the subsequent SFD procedure added an average of 58 seconds. These times were derived from 25 independent runs on the selected dataset using the same hardware for all computations. This additional $19\%$ of training time invested in feature selection is relatively small considering that SFD efficiently reduces model size by $98.86\%$ while providing an average increase in test accuracy of $0.6\%$.

\subsection{Code Availability}

The code used in this study is publicly available in the following GitHub repository: \href{https://github.com/gon-uri/detach_rocket}{\url{https://github.com/gon-uri/detach_rocket}}. It includes functions for end-to-end training of Detach-ROCKET for time series classification, as well as a standalone function for Sequential Feature Detachment. The latter is not limited to time series and is applicable to other domains where data have a large number of features and feature selection is required, such as bioinformatics and finance \cite{ma2008penalized,liang2015effect}.

\section{Discussion}
\label{sec:discussion}

\subsection{Comparison with Previous Methodologies}
\label{sec:discussion_comparison}

Comparing SFD with a standard backward-stepwise selection like SSR reveals the efficiency and adaptability of SFD, especially when dealing with large feature sets. SSR takes a step-by-step approach, removing features one at a time. This requires training a ridge classifier for each feature, which leads to computational bottlenecks, especially when the number of features ($f$) is large. SFD, on the other hand, uses a more granular strategy. It prunes several features in each iteration, with the number adaptively determined based on the remaining number of features. Early in the pruning process, when many features are still available, SFD can aggressively remove more of them. As the pool shrinks, SFD becomes more conservative, ensuring that it does not inadvertently remove an entire set of correlated features, which could adversely affect model performance.

Another notable characteristic of SFD is its reduced sensitivity to hyperparameter choices; the default value for the proportion of removed features ($p$) suffices in most cases. This contrasts with STLSQ, where the feature coefficient threshold is critical for the pruning result, and its selection can be very challenging, especially for randomly generated features.

In a direct comparison with alternative pruning strategies for the ROCKET model, S-ROCKET and POCKET, Detach-ROCKET provides equivalent classification accuracy, but with a significantly larger reduction in the number of features. Furthermore, unlike previous methods, Detach-ROCKET provides a way to automatically control the relationship between model size and model accuracy by varying the value of an interpretable parameter. 


\subsection{Interpretability and Theoretical Considerations.}

Selecting a subset of features not only provides an advantage in terms of accuracy, inference time, and model size, but also provides an opportunity for model interpretability. By comparing the properties of the initial set of kernels with the properties of the set of kernels retained after pruning (those associated with the retained features), it is possible to learn characteristics of the time series that are relevant to classification. For example, the distribution of dilation values in the selected kernels carries information about which are the relevant time scales in the problem. Furthermore, it is possible to analyze in which part of the time series the selected kernels are predominantly activated, indicating whether there are certain time periods that are more relevant for class determination. The idea of using pruning as a way to improve interpretability is further explored in a follow-up publication \cite{solana2024}.


From a theoretical viewpoint, our results can be contextualized within broader machine learning paradigms, such as neural network pruning and the lottery ticket hypothesis \cite{frankle2018lottery,hoefler2021sparsity}. The latter theory postulates the existence of a smaller, "winning" subnetwork within a larger neural network, analogous to our concept of "winning kernels". Our feature pruning methodology in Detach-ROCKET could be seen as an analog to finding these winning subnetworks, but in the realm of time series classification models. Furthermore, the remarkable robustness of SFD results let us conjecture that the winning kernels may incorporate useful inductive biases for the different classification tasks, as is the case for lottery ticket subnetworks initialized with random weights \cite{zhou2019deconstructing} (Fig. 6 pag. 7), and also may contain the maximum amount of information about the time series generative process, as can be seen in compressed data representations \cite{marsili2022quantifying}.

\subsection{Limitations and Future Work}

One limitation observed in our study is that the pruned model does not significantly outperform the full model in terms of accuracy. One of the reasons for this may be due to the already high performance of the full model on the standard UCR benchmark. Future investigations on more challenging datasets would be valuable to further evaluate the performance of the pruned models.

A salient aspect of random convolutional kernel models like ROCKET is their reliance on a simple linear classifier for training, which can limit the expressiveness of the model. This design choice is motivated by the need to avoid overfitting on the inherently large number of features, a number that is required due to the nonlearnable (random) nature of the kernels, as discussed in Section \ref{sec:background_rocket}. Our approach effectively prunes this feature space, retaining only the most informative kernels for classification. This reduction in dimensionality opens the door for future research to investigate the utility of fitting more complex classifiers to the selected feature set, as the risk of overfitting would be mitigated by the reduced feature count. 

\section{Conclusion}
\label{sec:conclusion}

In this study, we introduce Sequential Feature Detachment (SFD), a novel method for pruning random convolutional kernel-based time series classifiers such as ROCKET, MiniRocket, and MultiRocket. Our results show that Detach-ROCKET models, which are ROCKET models pruned using SFD, are able to improve the performance of ROCKET while significantly reducing the number of features.

Our methodology identifies the minimal set of discriminative kernels essential for high-accuracy classification, facilitating the construction of compact models that rival the performance of more complex counterparts. This shallow and light architecture not only enhances interpretability, but also offers computational advantages, particularly in resource-constrained environments, as it requires less memory and fewer convolutions for inference compared to traditional ROCKET models. Furthermore, the SFD procedure can be applied to select relevant features generated by other types of time series transformations, as we exemplify in our github repository with the features produced by the popular tsfresh Python package \cite{christ2018time}. These contributions advance the field's ongoing efforts to develop efficient and interpretable machine learning models for time series classification.

\section*{Acknowledgments}
This work was in part financially supported by Digital Futures. Authors would like to thank Yago J.S. Cagnacci for engaging in insightful discussions.

\bibliographystyle{unsrt}  
\bibliography{references}

\clearpage
\begin{appendices}

\renewcommand{\thefigure}{A\arabic{figure}}
\renewcommand{\thetable}{B\arabic{figure}}

\setcounter{figure}{0}    
\setcounter{table}{0}

\section{Kolmogorov-Smirnov Test Results}

\begin{figure}[ht]
  \centering
  \includegraphics[width=0.7\textwidth]{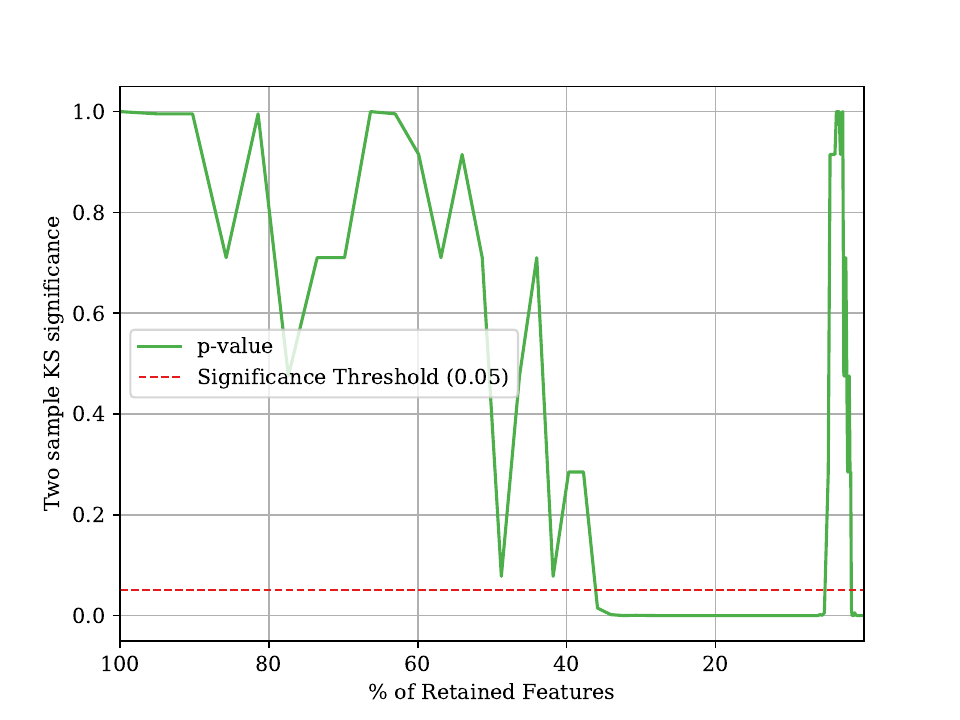}
  \caption{Results of the two-sample Kolmogorov-Smirnov test as a function of the percentage of features retained. At each pruning step, for the 25 experimental realizations, we test whether the number of pruned models that are better than the full model differs from the number of pruned models that are worse than the full model (violet and orange curves in Figure \ref{fig:fig1}). When feature retention is between $35.85\%$ and $5.10\%$, both distributions are significantly different because there are more better than worse models. When the feature retention is less than $1.73\%$, the distributions differ because there are significantly more worse models.}
  \label{fig:fig5}
\end{figure}

\clearpage

\section{Results Table}

\begin{table*}[ht]
    \centering
    \setlength{\tabcolsep}{2pt}
    \renewcommand{\arraystretch}{1.2}
    \begin{adjustbox}{width={\textwidth},totalheight={\textheight},keepaspectratio}%
 \begin{tabular}{l|cc|cc|cc}
    \hline

    & \multicolumn{2}{|c|}{Detach-ROCKET (10\%)} & \multicolumn{2}{|c|}{Detach-MiniRocket (10\%)} & \multicolumn{2}{|c}{Detach-MultiRocket (5\%)}\\
    
    
    
    Dataset (Train Size) & Test Acc. Full (\%)& Relative Change(\%) & Test Acc. Full (\%)& Relative Change(\%) & Test Acc. Full (\%)& Relative Change(\%)\\
    
    \midrule\midrule
    
    BeetleFly & 90.00$\pm$0.00 & -0.67$\pm$1.81 & 87.92$\pm$2.47 & -2.75 $\pm$ 3.65 & 85.00$\pm$0.00 & 3.53$\pm$2.88 \\
    BirdChicken & 90.00$\pm$0.00 & 0.00$\pm$0.00 & 90.00$\pm$0.00 & 0.00 $\pm$ 0.00 & 90.00$\pm$0.00 & 0.00$\pm$0.00 \\
    Chinatown & 98.20$\pm$0.11 & 0.23$\pm$0.19 & 98.23$\pm$0.08 & 0.02 $\pm$ 0.15 & 97.70$\pm$0.24 & 0.54$\pm$0.22 \\
    Coffee & 100.0$\pm$0.00 & 0.00$\pm$0.00 & 100.00$\pm$0.00 & 0.00 $\pm$ 0.00 & 100.00$\pm$0.00 & 0.00$\pm$0.00 \\
    Computers & 76.16$\pm$0.59 & 2.10$\pm$1.33 & 72.93$\pm$0.72 & -2.70 $\pm$ 2.20 & 79.00$\pm$0.60 & -2.43$\pm$0.74 \\
    DistalPhalanxOutlineCorrect & 77.22$\pm$0.58 & 0.79$\pm$1.09 & 78.17$\pm$0.85 & -1.07 $\pm$ 1.45 & 79.71$\pm$0.56 & -1.59$\pm$1.13 \\
    DodgerLoopGame & 84.66$\pm$0.59 & 1.23$\pm$1.05 & 85.30$\pm$0.37 & -0.31 $\pm$ 0.78 & 85.83$\pm$0.00 & 1.19$\pm$0.59 \\
    DodgerLoopWeekend & 98.41$\pm$0.00 & 0.00$\pm$0.00 & 98.41$\pm$0.00 & 0.00 $\pm$ 0.00 & 98.41$\pm$0.00 & 0.00$\pm$0.00 \\
    ECG200 & 90.32$\pm$0.47 & -2.13$\pm$0.53 & 91.75$\pm$0.43 & -1.00 $\pm$ 0.94 & 91.60$\pm$0.49 & -0.43$\pm$0.72 \\
    ECGFiveDays & 100.0$\pm$0.00 & 0.00$\pm$0.00 & 100.00$\pm$0.00 & 0.00 $\pm$ 0.00 & 100.00$\pm$0.00 & 0.00$\pm$0.00 \\
    Earthquakes & 74.82$\pm$0.00 & 0.00$\pm$0.00 & 74.82$\pm$0.00 & 1.52 $\pm$ 1.27 & 74.82$\pm$0.00 & -1.25$\pm$1.22 \\
    FordA & 94.46$\pm$0.29 & 0.31$\pm$0.36 & 95.09$\pm$0.25 & -0.48 $\pm$ 0.36 & 95.62$\pm$0.23 & -0.40$\pm$0.25 \\
    FordB & 80.59$\pm$0.54 & 0.17$\pm$0.93 & 81.85$\pm$0.39 & -0.75 $\pm$ 0.67 & 83.31$\pm$0.42 & -1.48$\pm$1.25 \\
    FreezerRegularTrain & 99.76$\pm$0.03 & 0.07$\pm$0.04 & 99.96$\pm$0.01 & 0.00 $\pm$ 0.02 & 99.96$\pm$0.00 & 0.00$\pm$0.00 \\
    FreezerSmallTrain & 95.26$\pm$0.55 & 0.99$\pm$0.33 & 96.84$\pm$0.38 & 0.64 $\pm$ 0.55 & 99.43$\pm$0.06 & -0.06$\pm$0.18 \\
    GunPoint & 100.0$\pm$0.00 & -0.03$\pm$0.13 & 99.33$\pm$0.00 & 0.22 $\pm$ 0.32 & 100.00$\pm$0.00 & 0.00$\pm$0.00 \\
    GunPointAgeSpan & 99.68$\pm$0.00 & 0.00$\pm$0.00 & 99.39$\pm$0.16 & -0.03 $\pm$ 0.16 & 100.00$\pm$0.00 & -0.03$\pm$0.09 \\
    GunPointMaleVersusFemale & 99.82$\pm$0.16 & 0.11$\pm$0.18 & 100.00$\pm$0.00 & 0.00 $\pm$ 0.00 & 100.00$\pm$0.00 & 0.00$\pm$0.00 \\
    GunPointOldVersusYoung & 99.12$\pm$0.21 & 0.05$\pm$0.23 & 100.00$\pm$0.00 & 0.00 $\pm$ 0.00 & 100.00$\pm$0.00 & 0.00$\pm$0.00 \\
    Ham & 72.53$\pm$1.54 & 1.81$\pm$2.22 & 71.11$\pm$1.47 & 1.73 $\pm$ 3.35 & 74.10$\pm$0.38 & 2.83$\pm$1.40 \\
    HandOutlines & 93.98$\pm$0.40 & -0.07$\pm$0.27 & 94.44$\pm$0.23 & -1.33 $\pm$ 0.93 & 94.76$\pm$0.18 & -0.20$\pm$0.51 \\
    Herring & 68.06$\pm$0.78 & -3.94$\pm$1.77 & 69.14$\pm$0.68 & -1.30 $\pm$ 3.36 & 68.44$\pm$0.94 & 9.38$\pm$2.45 \\
    HouseTwenty & 96.47$\pm$0.41 & -0.35$\pm$0.60 & 96.71$\pm$0.41 & 0.80 $\pm$ 0.43 & 98.66$\pm$0.56 & -0.34$\pm$0.57 \\
    ItalyPowerDemand & 96.93$\pm$0.09 & -0.14$\pm$0.16 & 96.62$\pm$0.16 & -0.23 $\pm$ 0.29 & 96.88$\pm$0.09 & 0.04$\pm$0.10 \\
    Lightning2 & 75.34$\pm$0.32 & 0.09$\pm$0.44 & 72.95$\pm$0.82 & 1.88 $\pm$ 0.84 & 68.69$\pm$0.49 & -1.19$\pm$1.60 \\
    MiddlePhalanxOutlineCorrect & 83.89$\pm$0.54 & 0.23$\pm$0.74 & 84.77$\pm$1.01 & -1.05 $\pm$ 2.55 & 85.43$\pm$0.83 & -0.92$\pm$0.82 \\
    MoteStrain & 91.55$\pm$0.28 & 0.17$\pm$0.28 & 92.76$\pm$0.24 & 0.20 $\pm$ 0.20 & 94.42$\pm$0.07 & 1.50$\pm$0.20 \\
    PhalangesOutlinesCorrect & 82.95$\pm$0.49 & -1.05$\pm$0.63 & 83.02$\pm$0.77 & -1.43 $\pm$ 1.81 & 84.94$\pm$0.35 & -1.41$\pm$0.54 \\
    PowerCons & 93.60$\pm$0.79 & 2.12$\pm$1.31 & 99.12$\pm$0.27 & -0.79 $\pm$ 1.03 & 98.67$\pm$0.51 & 0.00$\pm$0.71 \\
    ProximalPhalanxOutlineCorrect & 90.03$\pm$0.76 & -0.35$\pm$0.94 & 90.21$\pm$0.62 & -0.44 $\pm$ 0.67 & 92.03$\pm$0.61 & -0.11$\pm$0.41 \\
    SemgHandGenderCh2 & 92.36$\pm$0.76 & -0.71$\pm$0.71 & 89.71$\pm$1.26 & -4.98 $\pm$ 1.02 & 95.33$\pm$0.32 & 0.25$\pm$0.65 \\
    ShapeletSim & 99.98$\pm$0.11 & 0.02$\pm$0.11 & 100.00$\pm$0.00 & 0.00 $\pm$ 0.00 & 100.00$\pm$0.00 & 0.00$\pm$0.00 \\
    SonyAIBORobotSurface1 & 92.11$\pm$0.27 & 1.45$\pm$0.46 & 89.18$\pm$0.33 & 1.11 $\pm$ 0.62 & 88.45$\pm$0.19 & 3.14$\pm$0.24 \\
    SonyAIBORobotSurface2 & 91.42$\pm$0.20 & 0.57$\pm$0.40 & 91.53$\pm$0.22 & 1.17 $\pm$ 0.38 & 93.89$\pm$0.14 & 0.99$\pm$0.18 \\
    Strawberry & 98.26$\pm$0.22 & -0.15$\pm$0.26 & 98.27$\pm$0.13 & -0.16 $\pm$ 0.18 & 97.43$\pm$0.22 & -0.11$\pm$0.22 \\
    ToeSegmentation1 & 96.75$\pm$0.30 & -0.04$\pm$0.44 & 95.91$\pm$0.21 & -0.88 $\pm$ 0.51 & 95.26$\pm$0.18 & 1.24$\pm$0.36 \\
    ToeSegmentation2 & 92.31$\pm$0.92 & 0.17$\pm$0.92 & 91.86$\pm$0.38 & 0.49 $\pm$ 1.00 & 91.23$\pm$0.38 & 1.35$\pm$0.42 \\
    TwoLeadECG & 99.91$\pm$0.00 & 0.00$\pm$0.00 & 99.82$\pm$0.00 & 0.00 $\pm$ 0.00 & 99.82$\pm$0.00 & 0.04$\pm$0.04 \\
    Wafer & 99.83$\pm$0.03 & -0.01$\pm$0.04 & 99.92$\pm$0.02 & -0.01 $\pm$ 0.02 & 99.94$\pm$0.01 & -0.01$\pm$0.01 \\
    Wine & 81.19$\pm$2.95 & 2.35$\pm$4.02 & 85.34$\pm$2.33 & -1.95 $\pm$ 3.45 & 87.04$\pm$2.19 & 2.61$\pm$3.21 \\
    WormsTwoClass & 78.65$\pm$1.73 & 2.14$\pm$2.24 & 78.57$\pm$2.02 & -2.19 $\pm$ 2.33 & 78.57$\pm$0.65 & -0.83$\pm$1.84 \\
    Yoga & 91.15$\pm$0.35 & 0.10$\pm$0.36 & 90.75$\pm$0.20 & 0.11 $\pm$ 0.30 & 92.09$\pm$0.16 & -0.47$\pm$0.22 \\

    \midrule\midrule
    \textbf{AVERAGE} & 91.15 \ \ \ 0.35 & 0.18 \ \ \ 0.36 & 90.75
 \ \ \ 0.47 & -0.38 \ \ \ 0.90 & 91.34 \ \ \ 0.29 & 0.37 \ \ \ 0.62 \\
    \hline
\end{tabular}
\end{adjustbox}
    \caption{Results of the pruning process for ROCKET, MiniRocket and MultiRocket. We present the test accuracy of the Full Model on the left column, and the relative accuracy change after pruning on the right column. Results are reported as mean $\pm$ standard deviation. Experiments were repeted 25 times for ROCKET and 10 times for MiniRocket and MultiRocket.}
    \label{tab:detailed_results}
\end{table*}

\end{appendices}
\end{document}